# Dynamic consistency and decision making under vacuous belief

**Phan H. Giang**
George Mason University, MS 1J3
4400 University Dr., Fairfax, VA 22030, USA
*pgiang@gmu.edu*

## Abstract

The ideas about decision making under ignorance in economics are combined with the ideas about uncertainty representation in computer science. The combination sheds new light on the question of how artificial agents can act in a dynamically consistent manner. The notion of sequential consistency is formalized by adapting the law of iterated expectation for plausibility measures. The necessary and sufficient condition for a certainty equivalence operator for Nehring-Puppe's preference to be sequentially consistent is given. This result sheds light on the models of decision making under uncertainty.

## 1  INTRODUCTION

Ignorance as a state of knowledge is often characterized by the absence of relevant uncertainty information. Bayesian decision theory developed in the 1940s and 1950s, assuming the relevant risk is described by a probability function, ranks available actions by expected utility. The nature of the probability function can be objective (based on evidence) or subjective (based on personal preference). Almost immediately, many economists [19], [1] have been interested in the question how an individual should make decision if she can't associate any probability distribution to possible consequences of each alternative because of lack of evidence or violation of Savage's axioms. Following Savage's axiomatic approach, the earlier studies describe rational preference under ignorance by its characteristic properties.

The sustained interest in the problem is motivated by many real life circumstances in which ignorance arises naturally, for example, in competitive games where information about the opponent is scarce, unreliable, or even intentionally misleading. It is often a mistake to fill the void left by the lack of hard evidence with analytic and judgmental assumptions and then act upon those judgments as if they were facts.

In their attempts to create artificial agents with intelligence, computer scientists encounter the problem of how to program the behavior of the agents in the face of lack of information. While borrowing heavily from economists and philosophers, computer scientists also have to be concerned with the question of how to represent the state of ignorance. For economists, ignorance means the absence of probability information, therefore it can be declared by simply *leaving out* the probability component. For artificial agents, it is desirable to have a method for uncertainty representation in which the ignorance is obtained at the limit when the amount of information an agent has approaches zero. I call this representation *vacuous belief*. (This term is used here in its generic sense that includes but is not identical with a narrower usage in the Dempster-Shafer theory of belief functions.) In other words, "ignorance" and "vacuous belief" are synonyms but they differ by representation: the former is recognized by the absence of the uncertainty component, the latter is encoded by a particular representation of belief.

An important property of the expected utility (EU) model is the *dynamic consistency* which can be identified with the law of iterated expectation. $\mathbb{E}[U] = \mathbb{E}[\mathbb{E}[U|W]]$ where random variable $U$ represents utility of an act under risk and $W$ is information that can be observed. The expectation of $U$ can be calculated by, first, taking expectations of $U$ given the values of $W$ and then, taking expectation on the conditional expected values. The practical importance of the law is self-evident given its widespread application. In particular, neither the folding-back procedure nor the divide-and-conquer planning technique used to control complexity would be possible without this property.

In this paper, I examine decision making with vacuous belief, the representations of ignorance. Unlike most of the works on this topic, I insist that vacuous belief is just a state of belief. Hence, decision making under ignorance is decision making in which the state of belief happens to be vacuous. This view allows me to ask questions about dynamic consistency that are not sensible if ignorance is declared by the absence of the uncertainty component.

## 2 REPRESENTATION OF IGNORANCE

Assume a finite *state space* $\Omega$, the set of possible states of nature. The uncertainty information is encoded in the form of a *plausibility* function $\Delta$. Here I use the term in accordance with Halpern's definition [8], a representation of uncertainty that generalizes many familiar approaches such as probability theory, sets of probability measures, Dempster-Shafer belief function theory [20] and possibility theory [4]. It is convenient to think of states of $\Omega$ in terms of relevant *variables*. Variables are denoted by the capital letters to the end of the alphabet: $X, Y, Z$ whose values are denoted by lower case letters. A state is a tuple or concatenation of values, one for each variable. The terms *events* and *subsets of states* are used interchangeably and denoted by capital letters to the start of the alphabet: $A, B, C$.

$\Delta$ is a function from the algebra of subsets of $\Omega$, the *frame of reference*, to a partially ordered set $\mathcal{S}$, the *plausibility scale*. $\mathcal{S}$ includes the top ($\top$) and the bottom ($\bot$) elements and is equipped with relation $\geq$ that is reflexive, transitive and anti-symmetric. Conceptually, $\top$ means certainty, $\bot$ - impossibility and $\geq$ reads "more plausible". Three axioms are assumed for plausibility measure: $\Delta(\emptyset) = \bot$; $\Delta(\Omega) = \top$; and If $A \subseteq B$ then $\Delta(B) \geq \Delta(A)$.

$\Delta$ is equipped with two operators. The definition of the *restriction* operator is given below. The *conditioning* operator, given $\Delta$, creates a plausibility measure $\Delta_A$ on the algebra of subsets of $A$.

**Definition 1** *Suppose $\mathcal{H} = \{A_i\}_{i=1}^{m}$ is partition of $\Omega$ and $\Delta$ is a plausibility on the algebra of subsets of $\Omega$, the* restriction *of $\Delta$ on $\mathcal{H}$ is the mapping $\Delta^{\mathcal{H}}$ defined on the algebra based on partition $\mathcal{H}$ as follows: for any $I \subseteq \{1, 2, \ldots m\}$, $\Delta^{\mathcal{H}}(\cup_{i \in I} A_i) \mapsto \Delta(\cup_{i \in I} A_i)$.*

It is clear that $\Delta^{\mathcal{H}}$ satisfies all three plausibility axioms and hence is a plausibility function. Viewing each variable as a dimension one uses to describe the world, restriction obtains when one or more variables/dimensions are removed.

What plausibility function can be used for vacuous belief? In their ground-breaking work in early 1950s, Hurwicz & Arrow [1] argue that choice under ignorance should be invariant with respect to *relabeling* the states (property B in their paper) and deletion of *duplicate states* (property C). Define an act to be a mapping from $\Omega$ to a set of consequences. A state $\omega \in \Omega$ is called *duplicate* wrt a set of acts $D$ if there is another $\omega' \in \Omega$ and $\omega \neq \omega'$ such that $\forall f \in D, f(\omega) = f(\omega')$. Much later and in a different context of statistical inference, Walley [24] proposes two criteria that a vacuous belief must satisfy. The *embedding* principle requires that plausibility of an event $A$ should not depend on the sample space in which $A$ is embedded. The *symmetry* principle says that all elements in the sample space should be assigned the same plausibility. Intuitively, the symmetry principle and the relabeling invariance have the same informative content and so do the embedding principle and the deletion invariance. I formalize these criteria.

**Definition 2** *A class $\mathbf{V}$ of plausibility functions of finite domains on plausibility scale $\mathcal{S}$ is called* vacuous *if there exists $v \in \mathcal{S}$, called* vacuous element *such that for $\mathcal{IG}_{\Omega} \in \mathbf{V}$ and $\emptyset \neq A \subset \Omega$ then $\mathcal{IG}_{\Omega}(A) = v$.*

This definition is met by several intuitive representations of ignorance found in literature. The first example of vacuous belief is the set of all probability functions on $\Omega$. The plausibility scale is the set of pairs $\mathcal{Z} = \{\langle a, b \rangle \mid 0 \leq a \leq b \leq 1\}$ interpreted as lower and upper probabilities. The partial order $\langle a, b \rangle \geq_{\mathcal{Z}} \langle a', b' \rangle$ if $a \geq a'$ and $b \geq b'$. The vacuous element is $v = \langle 0, 1 \rangle$. For DS belief function theory, a belief function maps each event $A$ to a pair $\langle Bel(A), Pl(A) \rangle$. A vacuous belief function, according to the above definition, is one that has $\Omega$ as the only focus $m(\Omega) = 1$. In the possibility theory, vacuous belief is the function $\forall \omega \in \Omega, \pi(\omega) = 1$. Using the characterization of DS belief function by a convex set of probability functions and the characterization of a possibility function by a consonant belief function, one can show formally that those representations of ignorance are equivalent [20].

Several remarks are necessary. First, no single probability function satisfies the definition of vacuous belief. Second, the choice of plausibility scale $\mathcal{S}$ is important in meeting the definition. In particular, Halpern's proposal [8] (p.53) would not work. Suppose $\mathcal{P}$ is a set of probability measures on $\Omega$, Halpern suggests to use the set of all functions from $\mathcal{P}$ to $[0, 1]$ with $\geq_{\mathcal{S}}$ be the pointwise ordering on functions as the plausibility scale $\mathcal{S}$. That is, the plausibility value $Pl_{\mathcal{P}}(A)$ is a function $p_A$ such that $\forall \mu \in \mathcal{P}, p_A(\mu) = \mu(A)$. The advantage of using this plausibility scale in comparison with a more traditional way of taking lower and upper envelopes is that it can distinguish between different sets of probability measures that have the same lower and upper envelopes. The downside of this proposal is that it makes intuitively equal plausibility values incomparable. For example, consider the set of all probability measures, Halpern's plausibility scale would make any two events incomparable except when they are included in one another. In the lower-upper envelope approach all the events have the same plausibility.

From the definitions, a lemma follows immediately.

**Lemma 1** *The restriction of vacuous belief $\mathcal{IG}_\Omega$ on a partition $\mathcal{H} = \{A_i\}_{i=1}^k$ of $\Omega$ is also a vacuous belief.*

A necessary part of any description of a plausibility measure is the specification of a conditioning operator that is used to update belief upon the arrival of new evidence. There are different ways to define conditioning even for the same plausibility measure. Halpern [8] (Ch. 3) provides an excellent discussion of the issues. Despite the variety, as our interest in this case limits itself to vacuous belief, a minimal requirement for a conditioning operator to be considered reasonable is that conditioning of a vacuous belief on any event is again vacuous. Formally, I state it as an assumption.

**Assumption 1** *Suppose $\mathcal{IG}_\Omega$ is the vacuous belief on $\Omega$. For any $A \subset \Omega$, the conditional plausibility on $A$, $\mathcal{IG}_A$, is again vacuous.*

Note the convention by which conditioning is denoted by subscript and restricting by superscript. It can be seen that this assumption is satisfied by the conditioning defined for sets of probability measures; by both types of conditioning defined for DS belief functions (one is based on the interpretation of DS belief functions as sets of probabilities and the other is based on Dempster's rule of combination); and by both types of conditioning (quantitative and ordinal) defined for possibility functions [4]. The assumption follows from an intuitive argument. If one starts with vacuous belief, the events in $\Omega$ are indistinguishable in terms of plausibility. After observation of the truth of $A$, the events that are not compatible with $A$ are excluded. For the remaining events, both the symmetry and embedding principles still apply. Therefore, the conditional plausibility should be vacuous.

In sum, the class of vacuous plausibilities is closed under conditioning and restricting operators.

## 3  DECISION UNDER IGNORANCE

A decision situation can be described by a tuple $(\Omega, \Delta, \mathcal{U}, \mathcal{D})$. $\Omega$ and $\Delta$ are defined in section 2. $\mathcal{U} = [0,1]$ is the *outcome space*. The numeric value of an outcome is interpreted in a generic *utility* unit. Any bounded set of consequences that satisfies conditions of Debreu's theorem [3] can be converted to the unit interval. $\mathcal{D}$ is the *act space*. An *act* is a mapping $\Omega \to \mathcal{U}$. It is said to be *resolved* on a set of variables if knowing the values of the variables is sufficient to determine the outcome of the act. A *constant act* is $\forall s \in \Omega, f(s) = c$. Constant acts are denoted by their outcomes. An act can also be written as a contract consisting of a set of rules $\{A_i \hookrightarrow w_i | i = 1, m\}$ where $f^{-1}$ is a set-valued inverse of $f$ and $A_i = f^{-1}(w_i)$. Rule $A_i \hookrightarrow w_i$ means "if $A_i$ occurs then $w_i$ obtains". It is clear that the collection $\{A_i\}$ is a partition of $\Omega$. Two acts $f, h$, despite notational difference, are equivalent if they deliver the same outcome in each state.

A preference relation $\succeq$ on acts is an order i.e., is *complete* and *transitive* and *reflexive*. The asymmetric and symmetric parts of $\succeq$ are denoted by $\succ$ and $\sim$ respectively. $\succeq$ is required to reduce to the numerical order on constant acts i.e., for $c_1, c_2 \in \mathcal{U}$, $c_1 \succeq c_2$ iff $c_1 \geq c_2$.

Decision under ignorance has been discussed, among others, by Maskin [13], Nehring & Puppe [15] and more recently Puppe & Schlag [16] and Larbi et al [11]. Due to the lack of uncertainty information, it is a convention to identify an act with its set of possible outcomes. The ground-breaking result was made in early 1950s by Hurwicz & Arrow [1]. The basic construct is a *choice operator* ($\hat{\ }$) that for each set of available acts $D$ returns a subset of optimal acts $\hat{D} \subseteq D$. Arrow and Hurwicz postulated four rationality properties (properties A to D) that the choice operator must satisfy. They have shown that under complete ignorance, only extreme (the best and the worst) consequences matter. This result characterizes a family of utility functions including max, min and linear combinations of minimal and maximal values. Arrow-Hurwicz's conclusion has also been shown to hold under less stringent conditions. For example, in [11], it has been shown that the same characterization holds even without the construct of state space (acts are viewed as sets of consequences not as mappings from states to consequences).

Here I follow the approach by Nehring and Puppe [15]. Denote by $\mathcal{F}(\mathcal{U})$ the set of finite non-empty subsets of $\mathcal{U}$. The authors consider a *partial* order $\succeq_{NP}$ (transitive and reflexive) on $\mathcal{F}(\mathcal{U})$ and require the satisfaction of two conditions described below.

**Context independence (I)** Suppose $A \in \mathcal{F}(\mathcal{U})$, $x, y \in \mathcal{U}$ and $x, y \notin A$, if $x \geq y$ then $A \cup \{x\} \succeq_{NP} A \cup \{y\}$.

The definition of the continuity condition uses the concept of convergence of subsets in Hausdorff space.

**Definition 3 (Convergence on $\mathcal{F}(\mathcal{U})$)** *Denote $O_x$ for an open set containing $x$. A sequence $(A_i)_{i=1}^\infty$ in $\mathcal{F}(\mathcal{U})$ converges to $A \in \mathcal{F}(\mathcal{U})$ if for any collection of open sets $\mathcal{O} = \{O_x | x \in A\}$ and for all sufficiently large*

$n$: $A_n \subseteq \cup \mathcal{O}$ and $A_n \cap O_x \neq \emptyset$ for all $O_x \in \mathcal{O}$.

**Continuity (C)** Suppose $A, B \in \mathcal{F}(\mathcal{U})$ and sequence $(A_i)_{i=1}^{\infty}$ in $\mathcal{F}(\mathcal{U})$ converges to $A$. If $A_n \succeq_{NP} B$ for all $n$ then $A \succeq_{NP} B$

Condition (I) says that preference between two values by does not depend on the context i.e., the set of other outcomes they go with. Condition (C) requires $\succeq_{NP}$ to be continuous. Consider the set of ordered pairs $\mathcal{Z} = \{\langle a, b \rangle | 0 \leq a \leq b \leq 1\}$ and a preorder $\geqslant$ on $\mathcal{Z}$. Relation $\geqslant$ is said to be *continuous* if $\{y \in \mathcal{Z} | y \geqslant x\}$ and $\{y \in \mathcal{Z} | x \geqslant y\}$ are closed for any $x$. $\geqslant$ is said to be *weakly monotonic* if $a \geq a'$ and $b \geq b'$ implies $\langle a, b \rangle \geqslant \langle a', b' \rangle$. The main result of Nehring and Puppe is a theorem by which the preference between two sets of outcomes is decided by the preference between their pairs of minimal and maximal elements.

**Theorem 1 (Nehring and Puppe 1996)** $\succeq_{NP}$ on $\mathcal{F}(\mathcal{U})$ *satisfies conditions I and C iff there is a continuous and weakly monotonic $\geqslant$ on $\mathcal{Z}$ such that for $A, B \in \mathcal{F}(\mathcal{U})$, $A \succeq_{NP} B$ iff*

$$\langle \min(A), \max(A) \rangle \geqslant \langle \min(B), \max(B) \rangle \quad (1)$$

*Moreover, $\geqslant$ is unique. Conversely, given a continuous and weakly monotonic $\geqslant$ on $\mathcal{Z}$, $\succeq_{NP}$ on $\mathcal{F}(\mathcal{U})$ is uniquely determined.*

As mentioned before, in Nehring-Puppe's theorem $\succeq_{NP}$ is a partial order. For the rest of the paper, I assume that $\succeq_{NP}$ is complete and drop the subscript i.e., for any $A, B \in \mathcal{F}(\mathcal{U})$ either $A \succeq B$ or $B \succeq A$. By eq. 1, $\geqslant$ is also complete. Set $B = \{\min(A), \max(A)\}$ in eq. 1, it follows immediately that $A \sim \{\min(A), \max(A)\}$. For $z \in \mathcal{Z}$ define $z^{\uparrow} = \{x \in \mathcal{U} | \langle x, x \rangle \geqslant z\}$ and $z^{\downarrow} = \{x \in \mathcal{U} | z \geqslant \langle x, x \rangle\}$. Clearly, both $z^{\downarrow}$ and $z^{\uparrow}$ are nonempty ($0 \in z^{\downarrow}$ and $1 \in z^{\uparrow}$) and because of completeness of $\geqslant$, $z^{\downarrow} \cup z^{\uparrow} = \mathcal{U}$. By continuity, there is a unique $x \in \mathcal{U}$ such that $x \sim z$. A lemma follows.

**Lemma 2** *If $\succeq$ on $\mathcal{F}(\mathcal{U})$ is a weak order (complete, transitive and reflexive) and satisfies conditions I and C then for each $A \in \mathcal{F}(\mathcal{U})$ there is a unique $x \in \mathcal{U}$ such that $A \sim x$.*

Such a value is called the *certainty equivalence* or *ce* of $A$. Thus, lemma 2 establishes, from the semantics side, the *existence* of *ce* for any act (its outcomes).

From the operational side, the concern is to have a procedure that can *compute* the *ce* of any act under any plausibility measure. An *evaluation model* or *ce operator* is a functional $\mathcal{E} : \mathcal{C} \times \mathcal{D} \to \mathcal{U}$ where $\mathcal{C}$ is a class of plausibility measures and $\mathcal{D}$ is the set of acts. In general, $\mathbf{V} \subset \mathcal{C}$ i.e., the class of vacuous plausibilities is a subset of $\mathcal{C}$ but here I focus on $\mathbf{V}$ only. I say that $\mathcal{E}$ *correctly implements* $\succeq$ if

$$\mathcal{E}(\Delta, f) = x \text{ iff } f(\Omega) \sim x \quad (2)$$

where $\Delta$ is a vacuous plausibility and $f(\Omega)$ is the set of outcomes of $f$. This semantics implies several properties that $\mathcal{E}$ must satisfy.

**Unanimity** For $x \in \mathcal{U}$, $\mathcal{E}(\Delta, x) = x$.

**Range** $\mathcal{E}(\Delta, f) = \mathcal{E}(\Delta, \{\min(f(\Omega)), \max(f(\Omega))\})$.

**Monotonicity** If $a \geq a'$ and $b \geq b'$ then $\mathcal{E}(\Delta, \{a, b\}) \geq \mathcal{E}(\Delta, \{a', b'\})$.

**Continuity** $\lim_{x \to a} \mathcal{E}(\Delta, \{x, b\}) = \mathcal{E}(\Delta, \{a, b\})$ and $\lim_{x \to b} \mathcal{E}(\Delta, \{a, x\}) = \mathcal{E}(\Delta, \{a, b\})$.

The notation $\{a, b\}$ is used to denote any act $h$ such that $h(\Omega) = \{a, b\}$. Unanimity and Range are direct consequences of theorem 1. According to Unanimity, for constant acts, uncertainty does not matter. Monotonicity and Continuity of $\mathcal{E}$ follow from the weak monotonicity and the continuity of $\geqslant$.

These properties for $\mathcal{E}$ specify a wide class of functions that include familiar examples such as min, max and a weighted average of min and max (Hurwicz's $\alpha$-criterion). Max and Min functions can be characterized by imposing a new condition on $\succeq$. For example, the max function is characterized by (I), (C) and a new condition (M): if $A, B \in \mathcal{F}(\mathcal{U})$ and $A \subseteq B$ then $B \succeq A$. The Min function can be characterized by reversing the preference direction of (M) [15]. However, min and max are often criticized for being too extreme. In particular, (M) and its reverse are hard to defend based on common sense.

Hurwicz's criterion seems to be more reasonable but it has been found (for example in [9]) that it suffers from a "dynamic inconsistency" problem.

## 4 DYNAMIC CONSISTENCY OF DECISION WITH VACUOUS BELIEF

Dynamic consistency is a normative property that has a fundamental role in rational decision making. In the words of McClennen [14], it is about "consistency between planned choice and actual choice". It can also be viewed as a statement about decision maker's information processing efficiency. In this section, I'll be concerned with a specific form of dynamic consistency namely *sequential consistency* as formulated by Sarin and Wakker [17]. For computer scientists, this property is of particular importance because ubiquitous application of folding-back procedures would not be possible without it. I need some definitions leading to the formalization of sequential consistency.

**Definition 4** *For an act $f = \{A_i \hookrightarrow x_i\}_{i=1}^m$ and event $A \subset \Omega$, the conditional act $f_A$ is defined to be $\{A_i \cap A \hookrightarrow x_i | 1 \leq i \leq m, A_i \cap A \neq \emptyset\}$.*

**Definition 5** *If $f = \{E_i \hookrightarrow x_i\}_{i=1}^k$ then a new rule $A \hookrightarrow f$ is the set of rules obtained by prefixing $A \hookrightarrow$ to each rule in $f$ i.e., $A \hookrightarrow f = \{A \hookrightarrow E_i \hookrightarrow x_i\}_{i=1}^k$.*

The meaning of two-stage rule $A \hookrightarrow E_i \hookrightarrow x_i$ is that if $A$ is true *and* $E_i$ is true then the reward is $x_i$. This definition can be extended for multi-stage rules.

**Lemma 3** *For any partion $\mathcal{H} = \{A_k\}_{k=1}^K$ of $\Omega$, $f$ and $\{A_k \hookrightarrow f_{A_k}\}_{k=1}^K$ are equivalent.*

**Proof:** I have to show that for any $s \in \Omega$, $f(s)$ and the outcome received act $\{A_k \hookrightarrow f_{A_k}\}_{k=1}^K$ is the same. Suppose $f(\Omega) = \{w_j | 1 \leq j \leq J\}$, one can write the identity $f = \{f^{-1}(w_j) \hookrightarrow w_j\}_{j=1}^J$. So $f_{A_k} = \{A_k \cap f^{-1}(w_j) \hookrightarrow w_j\}_{j=1}^J$. Since $\{A_k\}_{k=1}^K$ is a partition of $\Omega$, $\cup_{k=1}^K (A_k \cap f^{-1}(w_j)) = f^{-1}(w_j)$. For any $s \in f^{-1}(w_j)$, there is $k$ such that $s \in A_k$. On the one hand $f(s) = w_j$. On the other hand, at $s$, $A_k$ is true and so is $A_k \cap f^{-1}(w_j)$, therefore act $\{A_k \hookrightarrow \{A_k \cap f^{-1}(w_j) \hookrightarrow w_j\}_{j=1}^J\}_{k=1}^K$ delivers $w_j$. ∎

I am now in the position to formalize the notion of sequential consistency.

**Definition 6** *Suppose $\mathcal{H} = \{A_i | 1 \leq i \leq m\}$ is a partition of $\Omega$. The ce operator $\mathcal{E}$ is said to be* sequentially consistent *wrt $\mathcal{H}$ if for any act $f$ and plausibility $\Delta$*

$$\mathcal{E}(\Delta, f) = \mathcal{E}(\Delta^{\mathcal{H}}, \{A_i \hookrightarrow \mathcal{E}(\Delta_{A_i}, f_{A_i})\}_{i=1}^m) \quad (3)$$

*where $\Delta^{\mathcal{H}}$ is the restriction of $\Delta$ on $\mathcal{H}$; $\{A_i \hookrightarrow \mathcal{E}(\Delta_{A_i}, f_{A_i})\}_{i=1}^m$ is an act defined on the algebra of $\mathcal{H}$; and $\Delta_{A_i}$, $f_{A_i}$ are conditionals of $\Delta$ and $f$ on $A_i$.*

This definition makes clear that the sequential consistency property is the basis of recursive (folding-back) evaluation of a decision tree. Partition $\mathcal{H}$ labels the branches emanating from a node attached with $\Delta$ and act $f$. Each of the children-nodes is attached with conditional plausibility measure $\Delta_{A_i}$ and conditional act $f_{A_i}$. $\mathcal{E}(\Delta, f)$ on the left-hand side of eq. 3 is the direct application of the *ce* operator for $\Delta$ and $f$. On the right-hand side, $\mathcal{E}(\Delta_{A_i}, f_{A_i})$ is the application of the *ce* operator for $i^{th}$ child-node. The obtained certainty equivalence replaces the subtree attached to that node i.e., the subtree is folded. The outer expression $\mathcal{E}(\Delta^{\mathcal{H}}, \{A_i \hookrightarrow \cdot\}_{i=1}^m)$ is the application of the *ce* operator for the remaining tree.

If $\Delta$ is a probability function ($P$), then eq. 3 is nothing but the law of iterated expectation where $\mathcal{E}(P, f)$ can be re-written in the familiar notation, $\mathbb{E}_P[f]$, and the right-hand side as $\mathbb{E}[\mathbb{E}[f|\mathcal{H}]]$.

In section 2, I argue that the vacuous plausibility class is closed under conditioning and restricting operations. That is if $\Delta \in \mathbf{V}$ then for any partition $\mathcal{H}$ and any event $A$, $\Delta^{\mathcal{H}} \in \mathbf{V}$ and $\Delta_A \in \mathbf{V}$. Thus, in eq. 3, all the applications of the *ce* operator are for vacuous belief.

I'll characterize a *ce* operator $\mathcal{E}$ for vacuous belief that satisfies all properties: Unanimity, Range, Monotonicity, Continuity and Sequential consistency. With the plausibility argument of $\mathcal{E}$ assumed to be vacuous, I can use a cleaner notation $\mathcal{E}_V : \mathcal{F}(\mathcal{U}) \to \mathcal{U}$. Consider a function $\gamma : \mathcal{Z} \to [0, 1]$ where, as before, $\mathcal{Z} = \{\langle a, b \rangle | 0 \leq a \leq b \leq 1\}$. Given the satisfaction of the Range property, $\mathcal{E}_V$ is completely determined by $\gamma$ and vice versa: $\forall A \in \mathcal{F}(\mathcal{U})$

$$\mathcal{E}_V(A) = x \Leftrightarrow \gamma(\min(A), \max(A)) = x \quad (4)$$

I'll translate the properties that $\mathcal{E}_V$ must satisfy into requirements for $\gamma$.

**Lemma 4** *Suppose functions $\mathcal{E}_V : \mathcal{F}(\mathcal{U}) \to \mathcal{U}$ and $\gamma : \mathcal{Z} \to [0, 1]$ are related via eq. 4. $\mathcal{E}_V$ satisfies Unanimity, Range, Monotonicity, Continuity and Sequential consistency iff $\gamma$ is continuous in each of the arguments and satisfies*
*(i) For $0 \leq x \leq 1$, $\gamma(x, x) = x$;*
*(ii) If $x \geq x', y \geq y'$ then $\gamma(x, y) \geq \gamma(x', y')$; and*
*(iii) For $0 \leq x \leq y \leq 1$*

$$\gamma(x, y) = \gamma(\gamma(x, x), \gamma(x, y)) = \gamma(\gamma(x, y), \gamma(y, y)) \quad (5)$$

**Proof:** (sketch) The facts that Range and Continuity of $\mathcal{E}_V$ are equivalent to $\gamma$ being continuous and determined by eq. 4 are clear. So is the correspondence between ($i$) and Unanimity; ($ii$) and Monotonicity. To see to the equivalence between Sequential consistency (eq. 3) and condition ($iii$). Choose an act $f$ such that $f(\Omega) = \{x, y\}$. Set $E = f^{-1}(x)$, (hence $f^{-1}(y) = \bar{E}$). Consider a partition $\mathcal{H} = \{A, \bar{A}\}$ such that $A \subset E$. Clearly, for conditional acts $f_A(A) = \{x\}$ and $f_{\bar{A}}(\bar{A}) = \{x, y\}$. Rewrite eq. 3 for $f$ and $\mathcal{H}$

$$\mathcal{E}_V(\{x, y\}) = \mathcal{E}_V(\{\mathcal{E}_V(x, x), \mathcal{E}_V(x, y)\}).$$

If one chooses $A$ for partition such that $A \supset E$ then

$$\mathcal{E}_V(\{x, y\}) = \mathcal{E}_V(\{\mathcal{E}_V(x, y), \mathcal{E}_V(y, y)\}).$$

Rewriting these equalities using eq. 4 yields eq. 5. ∎

Now I characterize function $\gamma$.

**Theorem 2** *Function $\gamma : \mathcal{Z} \to [0, 1]$ satisfies the conditions described in lemma 4 iff there exists a value*

$a \in [0,1]$ for which

$$\gamma(x,y) = \begin{cases} y & if\ y \leq a \\ a & if\ x \leq a \leq y \\ x & if\ x \geq a \end{cases} \quad (6)$$

**Lemma 5** *Suppose $\gamma$ satisfies properties (i), (ii) and (iii). If $\gamma(x,1) = a > x$ for some $x \in [0,1]$ then $\forall y \in [x,a]$, $\gamma(y,1) = a$. Similarly, if $\gamma(0,z) = a < z$ for some $z \in [0,1]$ then $\forall y \in [a,z]$, $\gamma(0,y) = a$.*

**Proof:** (lemma) By properties (i) and (iii), I have $a = \gamma(x,1) = \gamma(\gamma(x,x),\gamma(x,1)) = \gamma(x,a)$ and $a = \gamma(x,1) = \gamma(\gamma(x,1),\gamma(1,1)) = \gamma(a,1)$. It follows from property (ii) that for any $y$ in the interval $[x,a]$, $a = \gamma(x,a) \leq \gamma(y,1) \leq \gamma(a,1) = a$. ∎

**Proof:** (theorem) ($\Leftarrow$) It is necessary to show that the function $\gamma$ given by (6) is continuous on each argument and satisfy (i), (ii) and (iii). For a fixed value $x_0$, $\gamma(x_0,y)$ is continuous on $y$. In the case $x_0 \leq a$ and if $x_0 \leq b \leq a$ $\lim_{y \to b} \gamma(x_0,y) = b = \gamma(x_0,b)$. If $b \geq a$, $\lim_{y \to b} \gamma(x_0,y) = a = \gamma(x_0,b)$. In the case $x_0 \geq a$, $\lim_{y \to b} \gamma(x_0,y) = x_0 = \gamma(x_0,b)$. Similarly, it can be shown that $\gamma(x,y_0)$ is continuous on $x$ for any fixed $y_0$. The verification of (i) and (ii) is straightforward. Next, I have to show (iii) (eq. 5). In the first case, if $z \leq a$ then $\gamma(x,z) = z$, $\gamma(x,y) = y$ and $\gamma(y,z) = z$ hence $\gamma(\gamma(x,y),\gamma(y,z)) = \gamma(y,z) = z$. In the second case, if $x \geq a$ then $\gamma(x,z) = x$, $\gamma(x,y) = x$ and $\gamma(y,z) = y$, hence $\gamma(\gamma(x,y),\gamma(y,z)) = \gamma(x,y) = x$. In the third case $x \leq a$ and $z \geq a$, $\gamma(x,z) = a$. If $y \leq a$ then $\gamma(x,y) = y$ and $\gamma(y,z) = a$. Hence, $\gamma(\gamma(x,y),\gamma(y,z)) = \gamma(y,a) = a$. If $y \geq a$, $\gamma(x,y) = a$ and $\gamma(y,z) = y$, then $\gamma(\gamma(x,y),\gamma(y,z)) = a$.

($\Rightarrow$) Suppose $\gamma$ is a continuous function and satisfies (i), (ii) and (iii). Set $a = \gamma(0,1)$. Assume $0 < a < 1$. By lemma 5, $\gamma(0,a) = \gamma(a,1) = a$. By (ii), if $x \leq a \leq y$ then $a \leq \gamma(0,a) \leq \gamma(x,y) \leq \gamma(a,1) = a$. In other words, if arguments are in both sides of $a$ then $\gamma(x,y) = a$.

Now I consider the case when both arguments are on the same side of $a$. Assume that $x,y \geq a$. Our goal is to show by contradiction that $\gamma(x,y) = x$. Suppose, on the contrary, there exists $x_0, y_0 \geq a$ such that $\gamma(x_0,y_0) > x_0$. Because of (ii) $\gamma(x_0,1) = x_1 \geq \gamma(x_0,y_0) > x_0$. By lemma 5, $x$ in the interval $[x_0,x_1]$, $\gamma(x,1) = x_1$. Consider set $A_{x_1} = \{x | x \geq a, \gamma(x,1) = x_1\}$. This set includes interval $[x_0,x_1]$. Choose $a' = \inf A_{x_1}$. Clearly, $a \leq a' \leq x_0 < x_1$. For any $x$ in the neighborhood of $a'$ and $a' < x \leq x_1$, by definition of $a'$, there is $a' < x' \leq x$ such that $x' \in A_{x_1}$. By lemma 5, $\gamma(x,1) = x_1$.

I will show that for $x < a'$, $\gamma(x,1) \leq a'$. Suppose the contrary, $\gamma(x_3,1) = a'' > a'$ for some $x_3 < a'$. By lemma 5, for any $x \in [x_3,a'']$, $\gamma(x,1) = a''$. Clearly, the intersection $(a',x_1] \cap [x_3,a'']$ is not empty. For $x$ in the intersection, $\gamma(x,1) = x_1$ and $\gamma(x,1) = a''$. Hence $x_1 = a''$. Thus, for $x_3 < a'$ and $\gamma(x_3,1) = x_1$. This contradicts the fact that $a'$ is the infimum of $A_{x_1}$. So, $\gamma(x,1) \leq a'$ for $x < a'$.

For a sequence $\{x_i\}$ that approaches $a'$ from below $\lim_{x_i \uparrow a'} (\gamma(x_i,1)) \leq a'$. But for a sequence $\{y_i\}$ that approaches $a'$ from above $\lim_{y_i \downarrow a'} \gamma(y_i,1) = x_1$. So, $\gamma(\cdot,1)$ is not continuous at $a'$. This contradiction shows that $\gamma(x,y) = x$ for $a \leq x \leq y$. Similarly, one can show that $0 \leq x \leq y \leq a$, $\gamma(x,y) = y$. ∎

The main result combines lemma 4 and theorem 2.

**Theorem 3** *The ce operator $\mathcal{E}_V$ for vacuous belief satisfies properties Unanimity, Range, Monotonicity, Continuity and Sequential consistency iff there is a constant $a \in [0,1]$ such that for $A \in \mathcal{F}(\mathcal{U})$,*

$$\mathcal{E}_V(A) = \begin{cases} \max(A) & if\ \max(A) \leq a \\ a & if\ \min(A) \leq a \leq \max(A) \\ \min(A) & if\ \min(A) \geq a \end{cases} \quad (7)$$

**Corollary 1** *Suppose $\succeq$ is a weak order (reflexive, transitive and complete) that satisfies conditions (I) and (C). A sequentially consistent ce operator $\mathcal{E}_V$ correctly implements $\succeq$ iff it has the form (7).*

Form (7) includes min and max functions as special cases with $a = 0$ and $a = 1$ respectively. It is also clear that for any $0 < \alpha < 1$ Hurwicz's $\alpha$-criterion, $\alpha \min + (1-\alpha) \max$, does not satisfy (7).

The equality $a = \mathcal{E}_V(\{0,1\})$ gives an operational interpretation for the constant $a$. It is the value that the decision maker gives in exchange for set of outcomes $\{0,1\}$. The lower the value of $a$ the more uncertainty averse she is.

## 5 DISCUSSION

Enduring interest in the problem of decision making under ignorance can be explained, partly, by its puzzling nature, and partly, by the relevance to many practical situations (see [2] for a survey). Nehring and Puppe [15] characterizes preference $\succeq_\lambda$ that is complete and satisfies condition (C) and (SI) below.

**Strong independence (SI)** For all $A, B \in \mathcal{F}(\mathcal{U}), x \notin A \cup B$, if $A \succeq_\lambda B$ then $A \cup x \succeq_\lambda B \cup x$.

(SI) implies, hence, is strictly stronger than (I). For example both Hurwicz's $\alpha$-criterion with $\alpha \neq 0, 1$ and median rule ($A \succeq_{med} B$ iff $\text{med}(A) \geq \text{med}(B)$) satisfy (I) but not (SI). They show that there exists $\lambda \in [0,1]$ such that

$$A \succeq_\lambda B \text{ iff } \begin{cases} \lambda \geq \max(A) \geq \max(B)\ or \\ \min(A) \geq \min(B) \geq \lambda\ or \\ \max(A) \geq \lambda \geq \min(B) \end{cases} \quad (8)$$

It can be seen that the *ce* operator $\mathcal{E}_V$ in eq. 7 correctly implements $\succeq_\lambda$. Condition (I), although strictly weaker than (SI), in conjunction with (C) and Sequential consistency are equivalent to combination of (C) and (SI). In this relationship, Sequential consistency and (SI) reinforce each other.

The issue of dynamic inconsistency, the divergence between planned choice and actual choice, is raised in discussion of non expected utility models [14, 12, 17, 22]. The models that can account for observed systematic violations of SEU prescriptions, themselves violate the type of dynamic consistency that SEU satisfies. For an artificial agent, the dynamic consistency is important because it assures that the agent can achieve optimality by acting according to an optimal plan.

Although both decision making under ignorance and the dynamic consistency are under extensive discussion in the economics literature, to the best of my knowledge, the issues have never been examined in conjunction with each other. They seem to be incompatible. Being equalized to the absence of information, ignorance, like a ghost, does not have a representation. Under ignorance, acts are just sets of outcomes. On the other hand, dynamic consistency is all about updating and reaction to the updated information.

Computer scientists have many reasons to be interested in the notion of ignorance and its representation. Programming artificial agents, they have to treat ignorance differently by giving it a proper representation. The vacuous belief is just another state of belief on which conditioning and restricting operations are legitimate to perform. In order for a decision model to be dynamically consistent in general, it must be so in the case of the vacuous belief. Formula (7) provides a test case to disprove the sequential consistency claim.

More important value from the study of decision making with the vacuous belief comes from the fact that the vacuous belief is a common denominator on which different formal uncertainty frameworks converge. Any formal uncertainty framework worth its salt should be able to express *certainty* on one end of the spectrum and *ignorance* on the other end. Moreover, these concepts are clearly framework-independent. That is, their meaning does not depend on uncertainty language in which they are expressed.

As mentioned in section 2, the vacuous set of all probability measures, the vacuous DS belief function and the vacuous possibility function are all mathematically equivalent. So, at least they all pass the "worthy" test.

This makes the vacuous belief a unique testing ground in which decision models for various formal uncertainty frameworks can be directly compared. Unfortunately, this opportunity, so far, has not been exploited.

It is reasonable that different uncertainty frameworks necessitate different decision models: for belief represented by sets of probabilities [18, 23], for the DS belief function theory [10, 21, 7], and for possibility theory [5, 6]. However, the absence of a critical examination of the cottage industry of decision models contributes to their limited use in practice.

Here I outline a plan for such a comparative study. Suppose $(\Delta_i)_{i=1}^\infty$ is a sequence of plausibility measures converging to $\Delta$. The convergence $\lim_{i\to\infty} \Delta_i = \Delta$, can be suitably defined in the manner of definition 3. I can assume the continuity of the *ce* operator.

**Continuity on plausibility** For any converging sequence $(\Delta_i)_{i=1}^\infty$, $\lim_{\Delta_i \to \Delta} \mathcal{E}(\Delta_i, f) = \mathcal{E}(\Delta, f)$.

In particular, if $\Delta$ is vacuous, the condition becomes $\lim_{\Delta_i \to \Delta} \mathcal{E}(\Delta_i, f) = \mathcal{E}_V(f)$. That is, for a *ce* operator $\mathcal{E}$ to be sequentially consistent (for any plausibility measure), a necessary condition is that at the limit to the vacuous belief, it must have the form of eq. 7.

Just as all decision models must agree on an act under certainty, it is reasonable to expect that they do so when apply to the vacuous belief. In particular, denote by $\xi^{sp}, \xi^{ds}$ and $\xi^{po}$ the *representations of certainty* in three frameworks: the sets of probability functions, DS belief function theory and possibility theory and by $\omega^{sp}, \omega^{ds}$ and $\omega^{po}$ their *vacuous belief*.

**Consensus on certainty**

For any act $f$, $\mathcal{E}(\xi^{sp}, f) = \mathcal{E}(\xi^{ds}, f) = \mathcal{E}(\xi^{po}, f)$.

By itself, this equality is uninteresting and is trivially satisfied by all decision models. But the force of logic equally applies for the case of vacuous belief.

**Consensus on vacuous belief**

For any act $f$, $\mathcal{E}(\omega^{sp}, f) = \mathcal{E}(\omega^{ds}, f) = \mathcal{E}(\omega^{po}, f)$.

This requirement is nontrivial, in fact, has never been considered before in literature.

# 6 CONCLUSION

In this paper, I draw the ideas from research in economics on decision under ignorance and from research in computer science on uncertainty representation. Economists treat ignorance as a ghostlike object, "you can talk about it but you can't see it". For this concept, computer scientists have specific representations, the vacuous beliefs, and treat them like an ordinary state of belief. Because of that, the questions about dynamic consistency in decision making suddenly become sensible for the case of ignorance. I formalize the sequential consistency criterion by adapting the law of iterated expectation for plausibility measures. Based on Nehring-Puppe's characterization [15] of a preference under ignorance by Independence (I) and Continuity (C) conditions, I add a new requirement of sequential consistency. The necessary and sufficient condition for a sequentially consistent *ce* operator to correctly implement Nehring-Puppe's preference is given. I speculate about how this result can be used to compare the models of decision making under uncertainty.